# Using Machine Learning to Detect Fraudulent SMSs in Chichewa

Amelia Taylor [1][0000-0003-1485-8721] and Amoss Robert [2]

[1] Kuyesera AI Lab, Malawi University of Business and Applied Sciences, Blantyre, Malawi
ataylor@mubas.ac.mw

**Abstract** SMS enabled fraud is of great concern globally. Building classifiers based on machine learning for SMS fraud requires the use of suitable datasets for model training and validation. Most research has centred on the use of datasets of SMSs in English. Chichewa is a major language in Southern Africa, and is the language used most widely for communication in Malawi. This paper introduces a first dataset for SMS fraud detection in Chichewa and reports on experiments with machine learning algorithms for classifying SMSs in Chichewa as fraud or non-fraud. We answer the broader research question of how feasible it is to develop machine learning classification models for Chichewa SMSs. To do that, we created three datasets. A small dataset of SMS in Chichewa was collected through primary research from a segment of the young population in Blantyre city in Malawi. We applied a label-preserving text transformations to increase its size. The enlarged dataset was translated into English using two approaches: human translation and machine translation. The Chichewa and the translated datasets were subjected to machine classification using random forest and logistic regression. Our findings indicate that both models achieved a promising accuracy of over 96% on the Chichewa dataset. Since most machine learning models require data preprocessing, it is essential to investigate the impact of the reliance on English-specific tools for data preprocessing. We rerun the machine learning models on the English datasets obtained by translation. Our models had a drop in performance when moving from the Chichewa to the machine translated dataset. This highlights the importance of data preprocessing, especially in multilingual or cross-lingual NLP tasks, and shows the challenges of relying on machine-translated text for training machine learning models. Our results underscore the importance of developing language specific models for SMS fraud detection to optimise accuracy and performance.

**Keywords:** machine learning. SMS fraud, text augmentation, SMS classification.

## 1    Introduction

SMS services remain a critical part of telecommunications especially for people and places where access to internet remains challenging. SMS services are used for personal communication, notifications from various services such as banking, marketing campaigns. The key strength of SMS remains that it works independently of Internet connectivity. While the use of smartphones in Africa is on the rise, only 19% had access to internet on a smartphone in 2024 and SMSs remain an important medium of communication in rural and urban Africa [1]. However, the success of mobile technology globally and in Africa makes it prone to an increase in fraud. Globally, SMS enabled fraud resulted in large losses of $5.8 bn in 2023, a figure that increases every year as SMS fraud takes on new levels of sophistication [2].

SMS scams typically focus on getting people to share their personal data – this is called smishing. SMS spam is a general term used to refer to all unsolicited SMS messages, whether these contain advertising or are smishing. There is a significant increase in the prevalence and volume of SMS scams across African economies in recent years [3]. Bitdefender, a well-known cybersecurity and antivirus software company, discovered major scam attack campaigns in most countries ranging from package delivery to government related messages [4]. Africa is the second largest mobile market in the world after Asia. According to Bitdefender, South Africa, Ethiopia and Kenya, ranked as the top three countries in terms of the volume of spam SMS received per user. Data from smaller economies, such as Malawi, remains scarce.

Official statistics from Malawi are difficult to obtain. However, news reports often cite figures from the Malawi Communications Regulatory Authority (MACRA), which suggest that significant losses are attributed to SMS fraud involving mobile money transactions [5]. The latest public report of the Malawi Financial Intelligence Authority (FIA), entitled "Money Laundering Trends and Typologies", highlighted the fact that criminals made increasing use of SMSs, voice calls and Subscriber Identification Module (SIM) card swapping to engage in fraud and money laundering [6]. SMS fraud is typically linked to mobile money and other electronic payment services and SMS texts are a crucial part of the fraud cycle.

Unsuspecting victims are enticed through SMSs and calls to provide personal information to the fraudsters. Limited official information is available about the mechanisms through which SMS fraud occurs in Malawi and the success rate of the fraudsters. Occasional newspaper articles highlight individual cases, but detailed studies and publicly available datasets are scarce. The Money Laundering Trends and Typologies provides examples of SMS-enabled mobile money fraud in Malawi, particularly incidents that occurred during the COVID-19 period [6]. However, this report is not available to the public who is in most need of this information. In one example, scammers impersonated a legitimate NGO to defraud job seekers. They advertised positions for "COVID-19 preventive measures awareness officers," requiring applicants to pay a fee equivalent to $6 via a bank deposit or mobile money transfer. The scam was discovered when an applicant attempted to pay into the provided bank account rather than use mobile money, and this account was found to be incorrect although it bore some similarities to a real account for the organisation. The fraudsters relied on victims using the alternative mobile money wallet when the bank transaction failed. This scam involved

using a real NGO's name, fake bank account details, and a mobile wallet linked to another organisation to deceive desperate job seekers. Understanding the anatomy of an SMS fraud is important to help develop mechanisms in which fraud can be detected and for raising awareness among the general population who may be unsuspecting of the tactics that fraudsters employ.

Most techniques for SMS classification rely on textual features present in fraudulent SMSs, such as groups of words or special symbols, or frequent references to money. This implies the need to continuously develop datasets that are representative of the local contexts and languages. It has been recognised that collecting and maintaining legitimate SMSs is a challenging task due to privacy reasons [7]. Most researchers utilise datasets that are not publicly accessible or they use a combination of private and public datasets [8]. However, mixing data from different sources can lead to biased training since the corpus distribution might not reflect real cases [9].

In Malawi, most fraudulent SMSs that target the general population are written in Chichewa. Chichewa is a Bantu language with a wide distribution in Southern Africa. The language is the national language of Malawi and is also spoken in neighbouring countries such as Mozambique, Zambia, and Zimbabwe. Chichewa is considered a low-resourced language and has very few Natural Language Processing (NLP) tools available due to limited research and investment [10], [11], [12]. The lack of NLP tools is a common problem for many African languages. The scarcity of datasets in African languages exacerbates the challenges of applying traditional statistical methods to develop solutions for text classification [13].

Given the importance of understanding and preventing SMS fraud in general and specifically SMSs that use Chichewa, the aim of this research is:

   A. To construct a dataset of SMSs in Chichewa which can be used for understanding the anatomy of SMS-enabled fraud.
   B. To build and test machine learning classification algorithms that can effectively classify Chichewa SMS messages as fraud or normal.

In this paper we report on the preliminary results of this research.

### 1.1  Research on using machine learning for SMS fraud detection

Distinguishing fraudulent SMSs from normal ones is not a straightforward task as information about the nature of the SMS, whether fraudulent or not, is contained not only in the textual content of the SMS, but also in information about the sender [14]. When faced with a newly received SMS, a user relies on the textual content of the message and who the sender appears, to decide whether the SMS is fraudulent of not. Network provider can utilise this additional information contained in the header of an SMS to flag potentially dangerous messages and monitor traffic. Applications for filtering SMSs designed to run on smartphones could utilise both the content and the header of SMSs. However, ultimately the decision for the final classification of a newly received SMS remains with the user.

In this paper we look at SMS classification from the point of view of the user who relies on the text of the SMS to decide based on its content what the course of action would be. In this respect, SMS classification becomes a type of text classification which is an NLP task.

Text classification for SMSs has focused on two types of approaches: linguistic driven approaches and machine learning [14], [15]. Linguistic approaches focus on identifying textual patterns, linguistic clues and keywords in fraudulent messages that can be used to write rules. Linguistic approaches typically select a small number of most relevant features that describe fraudulent and normal SMSs. Fraudsters frequently change their tactics, and linguistic rules can quickly become outdated. Moreover, when it comes to under-resourced languages, linguistic tools such as stemming, named entity or parts of speech recognition, do not exist or are immature. This renders a purely linguistic approach to be inadequate for under-resources languages.

Machine learning (ML) approaches typically rely on the whole content of SMSs to construct relevant features that can be used for classification. The text is broken down into tokens / words which are 'encoded' numerically using word frequency techniques such as Bag-of-Words (BOW), TF-IDF (Term Frequency- Inverse Document Frequency), or more advanced vector embeddings such as word2vec [16], [17]. Each word becomes a feature and the space of features used by ML models is hyperdimensional. Numerical vector representations of words have succeeded in capturing fine-grained semantic and syntactic regularities using vector arithmetic despite the drawbacks of statistical semantics hypothesis [18]. This hypothesis relies on the assumption that statistical patterns of word usage is indicative of what people mean and word frequencies indicates their importance. So, if two words or two texts have the same vector or numerical representation then they are said to have similar meaning. Hybrid approaches combine text encodings and additional semantic labels constructed via linguistic means from the text [19].

Recent encodings using transformers are said to be more capable of capturing the meaning of words within their context without employing linguistic tools [20]. Transformers are the basis of powerful Large Language Models (LLMs) such as ChatGPT. To be useful transformers rely on deep neural network models trained on very large corpora. Once trained they offer deep contextualized word representations [21]. Linguistic and NLP tools are essential to developing the datasets needed for transformer training and fine tuning, but also for the evaluation, interpretability and application of their outputs [22].

Vector and transformer-based embeddings are available and well developed for the English language. For text in other languages there is a need to develop new embeddings. Such work is impaired by the availability of good quality and sufficiently large local text and by the intensive computation resources needed for deep training. Some under-resources languages have received more attention such as those for the Italian language, Arabic and languages spoken in India [23], [24], [25], [26]. In terms of SMS classification models for languages other than English, most have used simple word frequency techniques for embeddings such as BOW or TF-IDF or had to develop their own, usually small, custom embeddings [27], [28], [29], [30].

The first efforts for SMS spam filtering were focused on the English language, due to its prominence in global communication. Among these, the UCI SMS Spam Collection dataset has played a pivotal role. This dataset consists of thousands of labeled SMS messages categorised as either "ham" (legitimate) or "spam" (unsolicited) [31]. The UCI dataset became a benchmark for researchers in machine learning and natural language processing (NLP) and allowed researchers to evaluate and compare machine learning models, from classical methods like Naive Bayes and Support Vector Machines (SVMs) to modern deep learning techniques [31], [32]. While it provided a strong starting point, these efforts also highlighted challenges such as the need for more diverse datasets that include non-English languages, multilingual messages, and context-specific nuances.

Efforts are currently being made to expand this work to low-resource languages [9]. These studies typically focus on applying existing algorithms to small datasets sourced in local languages or datasets with mixed use of English and other languages. Recently, several authors explored methods for SMS fraud detection for languages of Indo, Turkic, Arabic [29], [33], [34], [35], [36]. Similar applications of machine learning have been used to classify SMS phishing messages in Swahili [30], [37], [38].

For African languages such as Chichewa, text is scarce, and NLP pipelines are patchy with some small efforts made for part of speech and name entity recognition on Chichewa text [12], [39]. Relying on embeddings of African text into English has been shown to be problematic [40].

The first ML experiments for SMS classification published in academic literature were done for the English language and employed traditional models such as Support Vector Machines and Naïve Bayes [14]. These models are considered benchmarks for SMS classification and have been revisited and improved over the years. Hybrid models combining multiple classifiers have demonstrated enhanced accuracy and precision, with one study reporting accuracies over 97% [41]. Experiments with deep learning techniques have recently been reported to surpass traditional models in terms of accuracy [42]. These models all show a better performance on non-spam SMSs, a comparatively low spam detection rate and high false positive rate. The 'concept drift problem' is one of the causes of this phenomena and it this refers to changes in wording used by fraudsters, resulting in poor performance on new unseen SMSs [42]. This makes the development of up-to date fraudulent SMSs imperative. Several recommendations have been put forward to help improve the state of the art of SMS fraud detection systems for the English language [42]. Two of the recommendation which we took on board in this research are (1) the use of crowdsourcing in collecting a good quality dataset of fraudulent SMSs and working on one-class detection (i.e. the class of fraudulent SMSs) and (2) the continuous experimentation and improvement of ML models on new and representative datasets.

## 2     Methodology

### 2.1     Data collection

We used crowdsourcing to collect SMSs from colleagues at the Malawi University of Business and Applied Sciences (MUBAS) who had been a target or a victim of SMS fraud. We approached the wider student community at MUBAS to contribute with additional SMSs for the project. Participants were asked to send SMSs that they considered to be fraudulent and SMSs they considered to be normal, and did not contain sensitive personal information. We used a snowball sampling approach by contacting those we knew had information about SMS fraud and we relied on their connections for additional SMSs. We also advertised via WhatsApp groups to seek participants that were willing to send us SMSs. SMSs were typically sent as screenshots or by direct forwarding to our devices. The data collection period was from March to August 2023. We also approached the local telecommunication companies and asked for fraudulent SMSs that had been officially reported by victims. A set of normal SMSs that contain service texts from telecommunication companies was also collected from users: this contained messages such as balance statement, and other service announcements.

Participants were not given any prior training on what constitutes a fraudulent SMS, but we responded to questions from participants by giving them explanations and examples of SMS fraud. To ensure data privacy, contributors were asked to share only non-private personal non-fraud messages, and others such as balance checks, bundle purchases, sports and religious news, music ads, promotions, fraud warnings from network operators.

### 2.2     Dataset construction

The received SMSs were entered manually in Microsoft Excel 2013. Source images were saved for reference, stored in a folder following a naming convention to help us match the data points in Excel to the source SMSs. The transcription of the SMSs included all typable characters in the SMS. We removed duplicates.

**Data augmentation** Dataset augmentation refers to techniques used to increase the diversity or size of a dataset by adding new data obtained from the original one by applying various transformations [43]. In our study, data augmentation was used to enlarge the size of the dataset. The transformations we applied on the text were meant to be label preserving, i.e. the meaning of the SMS was not changed.

**Table 1.** Data augmentation techniques.

| Augmentation technique | Description | Examples |
|---|---|---|
| Implicit meaning expansion | Add words implied by the context or meaning of the text but not explicitly stated. | Adding phrases such as: "*Mukudziwitsidwa kuti*" (English: '*you are informed that*') at the beginning of the text. |
| Synonym substitution | Replace words with their synonyms or equivalent expressions to maintain meaning. | The verb '*mulemele*' (en: '*be rich*') is replaced with the equivalent expression '*mupeze ndalama zankhani nkhani*' (en: '*to get a lot of money*'), which conveys the same imperative meaning. |
| Borrowed-to-vernacular translation | Replace borrowed words (e.g., from English) with their equivalents in the vernacular language. | The verb '*mujoine*' (borrowed from the English '*to join*') is replaced by the Chichewa '*kuti mulowe*' with the same meaning. |
| Morphological transformation | Alter the morphological structure of words, such as changing the suffix of a stem word. | We add the morpheme '*nd*' (en: '*am*') to the pronoun '*ine*' (en: '*I*'), and the morpheme '*mna*' (en: '*I was*') to the root verb '*patsidwa*' (en: '*given*') to reflect regional dialect variations. |

A fraudulent SMS did not become normal, and a normal SMS did not become fraudulent through data augmentation. The process of augmentation involved four transformations: (1) the addition of words that were implied by the text but were not explicitly present in the original Chichewa SMS; (2) replacing certain words with their synonyms or equivalent phrases; (3) replacing borrowed words (from English for example) with the equivalent words in vernacular (Chichewa) and (4) adding morphemes to some words. **Table 1** lists these transformations.

**Translation** We developed two additional datasets obtained via translation to English. One translation was done using Google Translate functions in Google Sheets and another was done manually by human translators. For the human translation we used three individual translators and conducted annotator agreement checks.

Therefore, in this paper we will work with three datasets: the Chichewa dataset (D-CHI), the human-translated dataset (D-HT), and the machine-translated dataset (D-MT).

**Annotation** Each SMS was labelled by contributors as being either fraudulent or normal. We conducted a second check on the labelling to avoid obvious mistakes. Messages such as balance checks, promotions although can be considered spam, in our work they were classified as non-fraudulent or normal. Label checking was done throughout the data collection. This enabled the researchers to learn a set of features that indicates fraud. Some of the labels given by participants were corrected in this process. We put together a list of common words or patterns commonly present in the fraudulent messages. The patterns were used to guide the labelling process of the dataset during the period of data collection.

### 2.3 Machine learning algorithms

We employed a supervised learning approach consisting of traditional machine learning models: logistic regression (LR), random forest (RF), support vector machine (SVM) and Naïve-Bayes (NB).

For feature extraction we used Term Frequency-Inverse Document Frequency (TF-IDF) to convert text into weighted numerical features. Due to the absence of vector embeddings for Chichewa we did not experiment with word2vec. To split text into tokens we used English stop words extended with a list of stop words in Chichewa built from our dataset. For all experiments, the data was split into 80% for training and 20% for testing.

Text classification algorithms, such SVM, and Naïve Bayes, have been frequently developed to build up search engines and construct spam filters. NB classifies documents based on the likelihood of a word occurring, utilising the Bayes theorem to calculate probabilities. Due to its simplicity and transparency, NB is a considered a benchmark model for text classification. Context information such as sentence structure, punctuation, order of words, or word relationships, such as word pairs, is not considered by this model. NB performs better when the classification classes are balanced. In our study, we start from a perfectly balanced dataset that contains the same number of fraudulent and normal SMSs.

SVM is a supervised learning algorithm that has often been used for pattern recognition. The algorithm classifies data by finding an optimal line or hyperplane that maximizes the distance between each class in an N-dimensional space. SVM can handle both linear and nonlinear classification tasks. When the data is not linearly separable, kernel functions are used to transform the data higher-dimensional space to enable linear separation. SVM is generally considered to outperform other methods for SMS fraud classification [31], [44].

LR continues to be one of the most used supervised learning techniques in data mining, especially for binary classification tasks [45]. The LR model requires the target variable to be binary, such as fraud or normal in this case and the independent variables to be numerical. LR is also highly interpretable, as each feature has a direct relationship with the log-odds of the outcome, making it valuable for understanding the influence of individual variables. Furthermore, while LR assumes a linear decision boundary, it can be adapted to handle challenges such as collinearity through regularisation.

RF is an ensemble learning method designed for both classification and regression tasks. It builds multiple decision trees during training and aggregates their outputs, using majority voting for classification [46]. The model leverages an ensemble of decision trees, where each tree is trained on random subsets of the data and features. By aggregating the outputs of these individual trees, RF improves robustness and accuracy in distinguishing fraudulent messages from legitimate ones. Random Forest can work well when combined with bag-of-words or TF-IDF features, especially if there is little need for capturing sequential dependencies. RF can handle high-dimensional data reasonably well, such as the TF-IDF features extracted from the SMS text and is effective at managing challenges like feature interactions and multicollinearity.

We applied Grid Search Hyperparameter Optimization (GSHO) for RF, SVM and NB [47]. Models were trained on three datasets: the Chichewa dataset (D-CHI), the human-translated dataset (D-HT), and the machine-translated dataset (D-MT). The models were then re-run on an extended datasets to check the sensitivity of the models in detecting fraudulent SMS in unbalanced datasets. To evaluate the models we used accuracy, precision, recall, F1-score and AUC. Results were compared with previous studies to validate the findings' relevance.

## 3    Results

### 3.1    The Main Dataset (D-CHI)

The main dataset collected through primary data collection contains SMSs in Chichewa. The initial dataset contained 101 fraudulent SMSs obtained from participants through crowdsourcing and a further 25 obtained from a telecommunication company. The dataset was enlarged through augmentation and the resulting dataset consisted of 676 SMS messages out of which half were labelled as fraudulent, and the other half were labelled as normal messages. Augmentation helped balance the fraud class with the normal class thereby improving class representation. **Table 1** shows the augmentation techniques we used. The resulting dataset is called D-CHI. We also created two additional datasets using human and machine translation: the human-translated dataset (D-HT), and the machine-translated dataset (D-MT).

We also use a dataset of 148 normal messages, telcoSMS, that contain service type messages sent by the telecommunication companies with balance statements and other announcements. A sample is given in **Table 2**. These messages contain balance statements, notification of service interruptions or new services and promotions. Usually, such messages tend to form the bulk of SMSs that subscribers receive. The datasets obtained thus by extension are denoted by D-CHIe, D-HTe, D-MTe – using the letter 'e' as a postfix.  In these extended datasets the ratio of fraudulent to normal SMS is 338:486 or 69% and the ratio of fraud to all SMSs is 40%. These extended datasets are used in our experiments to help explain the impact of class balance on ML models.

*Fig. 1* Extract from the main dataset in Chichewa.

| | SMS | Label |
|---|---|---|
| 1 | | |
| 2 | Mwagula Chezani PaNet MoFaya 3GB itha pa 05-11-2023 14:10:02 hrs. Imbani *137# kuti muo | normal |
| 3 | Sasangalalani! Mwalandila khetekhete bonasi ya mphindi 5 zoyimbira anzanu a Airtel. Imbar | normal |
| 4 | Tiketi imodzi YAULELE ya MK500 ikudikila inu 883155031.Muthakupata pompo. Tumizani YES | normal |
| 5 | TRANSPORTER:Ndine transporter kuchokela ku Joni ndanyamula katundu wa m'bale wanu n | fraud |
| 6 | Joyce Banda Foundation ikupeleka ndalama kwa azimayi zoyambila business. Tiyimbileni ph | fraud |
| 7 | Zosafunanso mayunitsi, mutha ku flasha anzanu ndi achibale kuma netiweki onse mwaulere | normal |
| 8 | tayimbaso apapa ndili pa line. | normal |

Table 2 Sample of normal SMS sent by telecommunication companies to customers indicating their balance or other services.

| Dataset | SMS ID | Text |
|---|---|---|
| D_CHI | TELCO0001 | Okondedwa akasitomala, mwatsala ndi MK-2199.20. Imbani *533# kuti mubwereke mayunitsi ndi KUTAPA. Sangalalani ndi macheza osatha ndi network ya Airtel. |
| D_HT | TELCO0001 | Dear customer, you have a balance of MK -2,199.20. Dial 533# to borrow airtime or data. Enjoy endless conversations with the Airtel network. |
| D_MT | TELCO0001 | Dear customer, you are left with MK-2199.20. Dial *533# to rent units with KATAPA. Enjoy endless chats with Airtel network. |

Table 3 Token statistics for the datasets showing the total number of tokens in each class and the average number of tokens per SMS.

| SMS Class | D-CHI, n | D-HT, n | D-MT, n | telcoSMS_CHI, n | telcoSMS_HT, n | telcoSMS_MT, n |
|---|---|---|---|---|---|---|
| Fraud | 6,001 | 7,211 | 6,826 | 0 | 0 | 0 |
| Normal | 3,956 | 4,927 | 4,780 | 3,530 | 4,019 | 3,785 |
| Total | 9,957 | 12,138 | 11,606 | 3,530 | 4,019 | 3,785 |
| Avg. any | 15 | 18 | 17 | 23 | 27 | 25 |
| Avg. Fraud | 18 | 21 | 20 | 0 | 0 | 0 |
| Avg. Normal | 12 | 15 | 14 | 24 | 27 | 26 |
| Unique Tokens | 2286 | 1580 | 2142 | 560 | 532 | 502 |

**Fig.** 1 shows an extract from the Chichewa dataset and **Table 2** shows examples of normal messages received from telecommunication companies. The size and number of tokens in the datasets is shown in **Table 3.** As observed in other datasets, fraudulent SMS typically contain more tokens than normal SMS. An observation also on the size of the MT versus HT: direct translation may be shorter but may fail to capture the exact

meaning of the original Chichewa text. As we can see the D-HT has more tokens than D-MT especially for the fraudulent SMSs.

### 3.2 Characteristics of the fraudulent SMSs

The top 10 most common expressions found in fraudulent SMSs are listed in **Table 4** and show distinct patterns. Some are linked to specific entities (e.g., JB Foundation and Mtukula Pakhomo), locations (e.g., South Africa and Johannesburg), and impersonation tactics (e.g., "ndine agent" and "transporter").

**Table 4.** Top 10 expressions appearing in the fraudulent SMSs.

| Chichewa expression | English translation |
|---|---|
| Ndine agent | I am an agent |
| South Africa | |
| Transporter | |
| Border | |
| Katundu | Luggage / goods |
| Mwachita mwayi | You have been lucky |
| Joni | Informal name for Johannesburg |
| JB Foundation | Joyce Banda (JB) Foundation supports impoverished women engaged in small businesses through anti-poverty grants. |
| Miracle Money | A term used in Malawi to refer to money obtained through some miracle or witchcraft. |
| Mtukula Pakhomo | Mtukula Pakhomo is a government-led social protection scheme that provides unconditional cash transfers to Malawi's poorest and most labour-constrained households. |

Our dataset shows that fraudsters exploit local knowledge and the vulnerabilities of specific segments of the population by referencing legitimate programs and initiatives to bait their victims. For instance, many Malawians have relatives working in South Africa often in cities like Johannesburg (commonly known as **Joni** in Malawi) and frequently receive packages through transporters. It is common for transporters to face delays at border crossings, such as the Mwanza border in Southern Malawi, while clearing passage into the country. Fraudsters leverage these cross-border challenges and the vulnerabilities of Malawian migrants in South Africa to deceive and steal from unsuspecting victims.

Extreme poverty in Malawi creates a fertile ground for deception, with more than 50% of people living below the poverty line [48]. Fraudsters exploit this vulnerability, using phrases like "miracle money" and "mwachita mwayi" ("you are lucky") to lure victims with promises of luck or supernatural financial gains. References to well-known programs and foundations offering cash assistance or support are often exploited in scams. Recognising these patterns enables machine learning models to effectively differentiate fraudulent SMS messages from legitimate ones, making them a vital tool against such schemes. Incorporating local knowledge is essential for these models, ensuring they are informed by recent developments and trends to maintain relevance and accuracy.

**Table 5** Characteristics of fraudulent SMSs *illustrating tactics used by scammers.*

| Characteristic | Description |
| --- | --- |
| Messages impersonating Government and NGO initiatives. | These messages impersonate programs such as social cash transfer schemes that provide financial support to poor families in Malawi. |
| SMSs impersonating mobile money agents. | These messages claim that money was accidentally sent to the subscriber and request that the amount should be returned. |
| SMSs impersonating transporters from South Africa. | These messages claim to be from a transporter who has carried a parcel from a subscriber's relative and requests money to clear it at the border. |
| SMSs impersonating mobile network operators or banking institutions. | These messages threaten subscribers with immediate account closure unless they follow specific instructions. |
| Messages inviting subscribers to join satanism. | These messages ask subscribers to send money to join satanism, with the promise of becoming rich. |
| Messages claiming the recipient won a prize. | These messages inform subscribers that they have won a prize and ask them to click on a link to redeem it, the subscriber must send money to them through the provided mobile money numbers. |

**Table 5** summarises the diverse range of tactics used by scammers to manipulate individuals into sending money or disclosing sensitive information, often using trusted entities or enticing promises as a means of deception. Understanding the characteristics of SMS fraud is vital for detecting, preventing, and responding to such scams effectively. Recognising common fraudulent patterns helps individuals and organisations identify suspicious messages, reducing the risk of financial loss. Fraud detection systems and machine learning models can also utilise these insights into to more accurately flag scam messages based on their characteristics.

## 3.3 Results of the machine learning experiments

**Summary of Models Accuracy**

RF and SVM models performed the best on all datasets while NB models had the worst performance. On the original dataset, D-CHI, Naïve Bayes and Logistic Regression performed worse than SVM and RF. We expected this to be the case given that LR and NB do not perform well when contextual understanding or the semantic relationships between words is important such as being the case here. SVM and RF deal better with high-dimensional and non-linear data, and SVM is more prone to overfitting. RF (best) on D-CHI gives an accuracy of 0.98, and an AUC-ROC of 0.99 and a FN rate of 3%, while SVM (best) has an accuracy of 0.96, AUC-ROC of 0.99 and a FN rate of 0%.

The accuracy numbers are quite high but reflect two important aspects of our data. Firstly, the dataset is balanced and contains a high quality of realistic fraudulent SMS and normal SMSs, making it easier for models to classify. Secondly, the dataset is small and any improvement in performance may come at the expense of overfitting.

**Fig. 2** The accuracy of the models on the three balanced datasets showing that RF has the highest performance on the D-CHI and lowest on the machine translated dataset. SVM has the highest performance on the D-HT. Among all models NB has the lowest performance on all datasets.

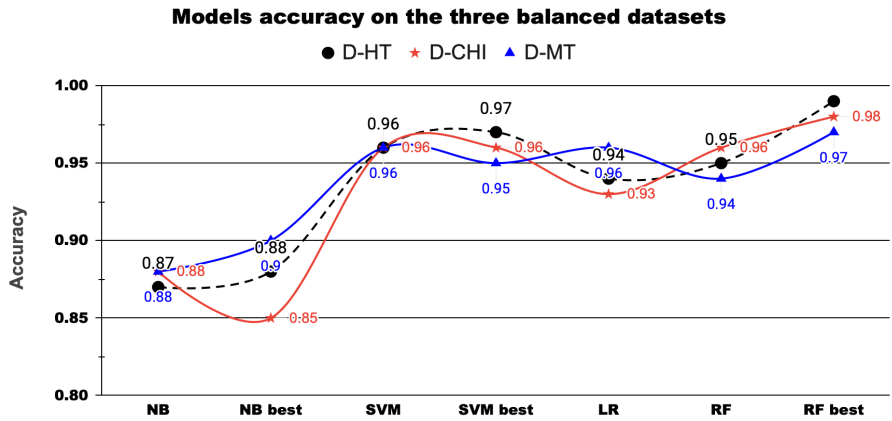

NB works well on simpler datasets, while SVM and RF often excel on more complex datasets. It is interesting to note that fine-tuned NB and LR performed worst on D-CHI compared to D-HT and D-MT. This can be explained by the limitation intrinsic to these models: feature independence, linearity, and poor performance on high-dimensional data. In our case, the feature space has on average 2000 features.

**Table 6** Accuracy of the machine learning experiments on the three balanced datasets and on the three extended datasets with normal SMSs. RF is the best performing model on the original D-CHI dataset. The colour codes are all comparing values against the values for D-CHI. Green represent an improvement and red a deterioration in a score. The results show that D-HT performs better in terms of all metrics for most of the models except for RF. D-MT leads to a lower performance compared to the D-CHI. Extending the datasets with more normal messages leads to an improvement in performance mainly for the original D-CHI dataset.

| | | | Fraud SMS | | | Normal SMS | | | | | | | | |
|---|---|---|---|---|---|---|---|---|---|---|---|---|---|---|
| | | Accuracy | Precision | Recall | F1-score | Precision | Recall | F1-score | AUC-ROC | Incorrect Classified | FN, n | FP, n | FP % | FN % |
| D-CHI | NB | 0.88 | 0.86 | 0.92 | 0.89 | 0.90 | 0.83 | 0.98 | 0.87 | 18 | 6 | 11 | 17 | 8 |
| | NB best | 0.85 | 0.80 | 0.96 | 0.88 | 0.94 | 0.73 | 0.82 | 0.98 | 20 | 3 | 17 | 27 | 4 |
| | SVM | 0.96 | 0.95 | 0.97 | 0.96 | 0.97 | 0.94 | 0.95 | 0.99 | 6 | 2 | 4 | 6 | 3 |
| | SVM best | 0.96 | 0.94 | 1.00 | 0.97 | 1.00 | 0.92 | 0.96 | 0.99 | 5 | 0 | 5 | 8 | 0 |
| | LR | 0.93 | 0.92 | 0.96 | 0.94 | 0.95 | 0.90 | 0.93 | 0.98 | 9 | 3 | 6 | 10 | 4 |
| | RF | 0.96 | 0.95 | 0.97 | 0.96 | 0.97 | 0.94 | 0.95 | 0.99 | 6 | 2 | 4 | 6 | 3 |
| | RF best | 0.98 | 0.99 | 0.97 | 0.98 | 0.97 | 0.98 | 0.98 | 0.99 | 3 | 2 | 1 | 2 | 3 |
| D-HT | NB | 0.87 | 0.86 | 0.90 | 0.88 | 0.88 | 0.83 | 0.85 | 0.86 | 18 | 7 | 11 | 17 | 10 |
| | NB best | 0.88 | 0.83 | 0.97 | 0.89 | 0.96 | 0.76 | 0.85 | 0.88 | 17 | 2 | 15 | 24 | 2 |
| | SVM | 0.96 | 0.95 | 0.97 | 0.96 | 0.97 | 0.94 | 0.95 | 0.996 | 6 | 2 | 4 | 6 | 3 |
| | SVM best | 0.97 | 0.95 | 1.00 | 0.97 | 1.00 | 0.94 | 0.97 | 0.99 | 4 | 0 | 4 | 6 | 0 |
| | LR | 0.94 | 0.93 | 0.96 | 0.95 | 0.95 | 0.92 | 0.94 | 0.99 | 8 | 3 | 5 | 8 | 4 |
| | RF | 0.95 | 0.92 | 0.99 | 0.95 | 0.98 | 0.90 | 0.94 | 0.99 | 7 | 1 | 6 | 10 | 1 |
| | RF best | 0.99 | 0.97 | 1.00 | 0.99 | 1.00 | 0.97 | 0.98 | 0.996 | 2 | 2 | 1 | 3 | 0 |
| D-MT | NB | 0.88 | 0.83 | 0.97 | 0.89 | 0.96 | 0.76 | 0.85 | 0.87 | 17 | 2 | 15 | 24 | 2 |
| | NB best | 0.90 | 0.88 | 0.96 | 0.92 | 0.95 | 0.84 | 0.89 | 0.94 | 13 | 3 | 10 | 16 | 4 |
| | SVM | 0.96 | 0.95 | 0.97 | 0.96 | 0.97 | 0.94 | 0.95 | 0.98 | 6 | 2 | 4 | 6 | 3 |
| | SVM best | 0.95 | 0.92 | 0.99 | 0.95 | 0.98 | 0.99 | 0.94 | 0.98 | 7 | 1 | 6 | 6 | 0 |
| | LR | 0.96 | 0.95 | 0.97 | 0.96 | 0.97 | 0.94 | 0.95 | 0.97 | 6 | 2 | 4 | 6 | 3 |
| | RF | 0.94 | 0.95 | 0.95 | 0.95 | 0.94 | 0.94 | 0.94 | 0.98 | 8 | 4 | 4 | 6 | 5 |
| | RF best | 0.97 | 0.97 | 0.97 | 0.97 | 0.97 | 0.97 | 0.97 | 0.998 | 4 | 2 | 2 | 3 | 3 |

| | | Accuracy | Fraud SMS | | | Normal SMS | | | AUC-ROC | Incorrect Classified | FN | FP | FP % | FN % |
|---|---|---|---|---|---|---|---|---|---|---|---|---|---|---|
| | | | Precision | Recall | F1-score | Precision | Recall | F1-score | | | | | | |
| D-CHI e | NB | 0.95 | 0.90 | 0.98 | 0.94 | 0.99 | 0.93 | 0.96 | 0.96 | 8 | 1 | 7 | 7 | 2 |
| | NB best | 0.91 | 0.87 | 0.94 | 0.91 | 0.95 | 0.89 | 0.92 | 0.92 | 16 | 2 | 14 | 14 | 3 |
| | SVM | 0.96 | 0.97 | 0.92 | 0.94 | 0.95 | 0.98 | 0.97 | 0.997 | 7 | 5 | 2 | 2 | 8 |
| | SVM best | 0.98 | 0.97 | 0.97 | 0.97 | 0.98 | 0.98 | 0.98 | 0.997 | 6 | 4 | 2 | 2 | 2 |
| | LR | 0.96 | 0.98 | 0.91 | 0.94 | 0.94 | 0.99 | 0.97 | 0.995 | 7 | 6 | 1 | 1 | 9 |
| | RF | 0.96 | 0.98 | 0.91 | 0.94 | 0.94 | 0.99 | 0.97 | 0.99 | 7 | 6 | 1 | 1 | 9 |
| | RF best | 0.96 | 0.95 | 0.94 | 0.94 | 0.96 | 0.97 | 0.97 | 0.997 | 7 | 4 | 3 | 3 | 6 |
| D-HT e | NB | 0.90 | 0.81 | 0.95 | 0.88 | 0.97 | 0.86 | 0.91 | 0.91 | 17 | 3 | 14 | 14 | 5 |
| | NB best | 0.89 | 0.80 | 0.95 | 0.87 | 0.97 | 0.85 | 0.91 | 0.90 | 17 | 3 | 15 | 15 | 5 |
| | SVM | 0.96 | 0.94 | 0.97 | 0.95 | 0.98 | 0.96 | 0.97 | 0.996 | 6 | 4 | 2 | 4 | 3 |
| | SVM best | 0.96 | 0.94 | 0.97 | 0.95 | 0.98 | 0.96 | 0.97 | 0.997 | 6 | 2 | 4 | 4 | 3 |
| | LR | 0.96 | 0.95 | 0.94 | 0.94 | 0.96 | 0.97 | 0.97 | 0.99 | 7 | 3 | 4 | 3 | 6 |
| | RF | 0.95 | 0.94 | 0.92 | 0.93 | 0.95 | 0.96 | 0.96 | 0.99 | 9 | 4 | 5 | 4 | 8 |
| | RF best | 0.95 | 0.97 | 0.91 | 0.94 | 0.94 | 0.98 | 0.96 | 0.99 | 8 | 2 | 6 | 2 | 9 |
| D-MT e | NB | 0.85 | 0.77 | 0.89 | 0.83 | 0.92 | 0.83 | 0.88 | 0.86 | 24 | 7 | 17 | 17 | 11 |
| | NB best | 0.89 | 0.85 | 0.88 | 0.86 | 0.92 | 0.90 | 0.91 | 0.94 | 18 | 8 | 10 | 10 | 13 |
| | SVM | 0.93 | 0.92 | 0.89 | 0.90 | 0.93 | 0.95 | 0.94 | 0.99 | 12 | 7 | 5 | 5 | 11 |
| | SVM best | 0.94 | 0.90 | 0.95 | 0.92 | 0.97 | 0.93 | 0.95 | 0.94 | 12 | 5 | 7 | 5 | 7 |
| | LR | 0.93 | 0.93 | 0.89 | 0.91 | 0.93 | 0.96 | 0.95 | 0.98 | 11 | 7 | 4 | 4 | 11 |
| | RF | 0.92 | 0.98 | 0.81 | 0.89 | 0.89 | 0.99 | 0.94 | 0.97 | 13 | 12 | 1 | 1 | 19 |
| | RF best | 0.92 | 0.92 | 0.88 | 0.90 | 0.92 | 0.95 | 0.94 | 0.98 | 13 | 8 | 5 | 5 | 13 |

LR often acts as a baseline for classification. Using LR can serve as a useful benchmark and help us see the relationship between the features (word frequencies or TF-IDF values) and the classification output. As the results show, when run on the balanced datasets D-CHI, D-HT, D-MT, the following order on performance is maintained NB <

LR < (SVM +RF). This reflects the bias-variance tradeoff observed in other ML studies:

> NB (High Bias, Low Variance, Easy to interpret) → LR (Balanced Bias-Variance, Easy to interpret) → (SVM, RF, More challenging to interpret) (Low Bias, Higher Variance, More challenging to interpret. )

**Table 6** gives an in-depth breakdown of the performance of the models in all experiments.

**Hyper-parameter tuning**
For SVM, NB and RF we run hyperparameter tuning using a Grid Search Optimisation which tests all combinations of the possible hyperparameters. For SVM and NB we used random stratified K-Fold optimisation. The optimised models did not necessarily result in a better performance on the test data due to the small dataset and the fact that they tended to overfit the training data. SVM is known to overfit when there is noise in high-dimensional text data. SMS datasets often have short text messages, which limits the amount of information each sample contains. Solutions to overfitting are related to increasing the size our dataset, use embeddings such as word2vec and deep learning models and use cross-validation techniques such as stratified k-fold. As SVM is more prone to overfitting we are experimenting with using stratified k-fold for SVM and enlarging the dataset using additional normal messages. Noting that the rate of FP is worse for all models across all datasets, enlarging with additional normal SMS should have an effect to reduce the FP rate.

**The effects of language**
It is expected that the human translation is more carefully done and uses consistent language across the two classes and that the models would perform quite similarly for D-CHI and D-HT and would be close in terms of levels of accuracy. As illustrated by **Fig. 2** there are large differences between the accuracy of models on D-CHI and D-MT, suggesting that machine translation (D-MT) had a larger negative impact on performance compared to human translation (D-HT). This difference in performance might be due to the potential loss of context or translation errors in the machine-translated dataset, which affects the model's ability to correctly classify the SMS messages.

Only NB and LR perform better on D-MT compared to the other two datasets, D-CHI and D-HT. The gap between the performance of NB on D-CHI and on the translated dataset if the largest of all the models. SVM un-tuned had an almost equal performance of all the datasets, and SVM best performed better on D-CHI and D-HT than on D-MT. Similarly for RF there is a decrease in performance when moving from the original dataset (D-CHI) to the translated datasets with the worse performance recorded for the machine-translated dataset (D-MT). However, RF models remain the better performers indicating that they might be better suited for handling the noise or inconsistencies introduced by the translation process.

*Table 7 Best hyperparameters for SVM, NB and RF.*

| Alg. | Dataset | Best Hyperparameters | Best Score |
|---|---|---|---|
| SVM | D-CHI | {'tol': 1e-6, 'kernel': 'linear', 'gamma': 0.1, 'C': 1000} | 0.97 |
| | D-CHIe | {'tol': 0.001, 'kernel': 'sigmoid', 'gamma': 1, 'C': 1000} | 0.97 |
| | D-HT | {'tol': 0.001, 'kernel': 'rfb', 'gamma': 0.1, 'C': 10} | 0.97 |
| | D-HTe | {'tol': 0.0001, 'kernel': 'linear', 'gamma': 1, 'C': 1} | 0.98 |
| | D-MT | {'tol': 0.001, 'kernel': 'rbf', 'gamma': 0.001, 'C': 1000} | 0.92 |
| | D-MTe | : {'tol': 0.001, 'kernel': 'rbf', 'gamma': 0.001, 'C': 1000} | 0.94 |
| RF | D-CHI | {'n_estimators': 180, 'min_samples_split': 5, 'min_samples_leaf': 1, 'max_features': 1, 'max_depth': 110, 'bootstrap': True} | 0.97 |
| | D-CHIe | {'n_estimators': 180, 'min_samples_split': 10, 'min_samples_leaf': 1, 'max_features': 1, 'max_depth': None, 'bootstrap': False} | 0.98 |
| | D-HT | {'n_estimators': 180, 'min_samples_split': 10, 'min_samples_leaf': 1, 'max_features': 1, 'max_depth': None, 'bootstrap': False} | 0.97 |
| | D-HTe | {'n_estimators': 230, 'min_samples_split': 10, 'min_samples_leaf': 1, 'max_features': 1, 'max_depth': 80, 'bootstrap': True} | 0.98 |
| | D-MT | {'n_estimators': 180, 'min_samples_split': 10, 'min_samples_leaf': 1, 'max_features': 1, 'max_depth': None, 'bootstrap': False} | 0.93 |
| | D-MTe | {'n_estimators': 180, 'min_samples_split': 10, 'min_samples_leaf': 1, 'max_features': 1, 'max_depth': None, 'bootstrap': False} | 0.94 |
| NB | D-CHI | {'var_smoothing': 0.123328467394420659} | 0.88 |
| | D-CHIe | {'var_smoothing': 1.0} | 0.89 |
| | D-HT | {'var_smoothing': 0.18738817422860384} | 0.85 |
| | D-HTe | {'var_smoothing': 0.0053366699231206307} | 0.87 |
| | D-MT | {'var_smoothing': 1.0} | 0.86 |
| | D-MTe | {'var_smoothing': 1.0} | 0.87 |

**The effect of class balance and SMS content**

To understand the impact of class balance and the content of the SMS, we run the same models on extended datasets obtained by adding normal SMSs, namely, D-CHIe, D-HTe, D-MTe. The performance of all models on D-CHIe improves except for the fine-tuned RF whose performance on fraudulent SMSs deteriorates (**Table 6**). The models trained on extended datasets are now better at classifying normal SMSs (have a lower FP rate) but are not better at detecting fraudulent SMSs.

*Table 8 False Positive (FP) and False Negative(FN) rates for the six datasets and 7 models. Highlighted are rates less than 5%..*

|  |  | NB | NB best | SVM | SVM best | LR | RF | RF best |
|---|---|---|---|---|---|---|---|---|
| D-CHI | FP, % | 17 | 27 | 6 | 8 | 10 | 6 | 2 |
|  | FN, % | 8 | 4 | 3 | 0 | 4 | 3 | 3 |
| D-HT | FP, % | 17 | 24 | 6 | 6 | 8 | 10 | 3 |
|  | FN, % | 10 | 2 | 3 | 0 | 4 | 1 | 0 |
| D-MT | FP, % | 24 | 16 | 6 | 6 | 6 | 6 | 3 |
|  | FN, % | 2 | 4 | 3 | 0 | 3 | 5 | 3 |
| D-CHIe | FP, % | 7 | 14 | 2 | 2 | 1 | 1 | 3 |
|  | FN, % | 2 | 3 | 8 | 2 | 9 | 9 | 6 |
| D-HTe | FP, % | 14 | 15 | 4 | 4 | 3 | 4 | 2 |
|  | FN, % | 5 | 5 | 3 | 3 | 6 | 8 | 9 |
| D-MTe | FP, % | 17 | 10 | 5 | 5 | 4 | 1 | 5 |
|  | FN, % | 11 | 13 | 11 | 7 | 11 | 19 | 13 |

*Fig. 3 Performance of models on the balanced and extended datasets. Extending the datasets with additional normal messages results in improved performance of most models on D-CHI but not so much for the translated datasets. Additional normal messages during training results in a better FP rate but a worse FN rate.*

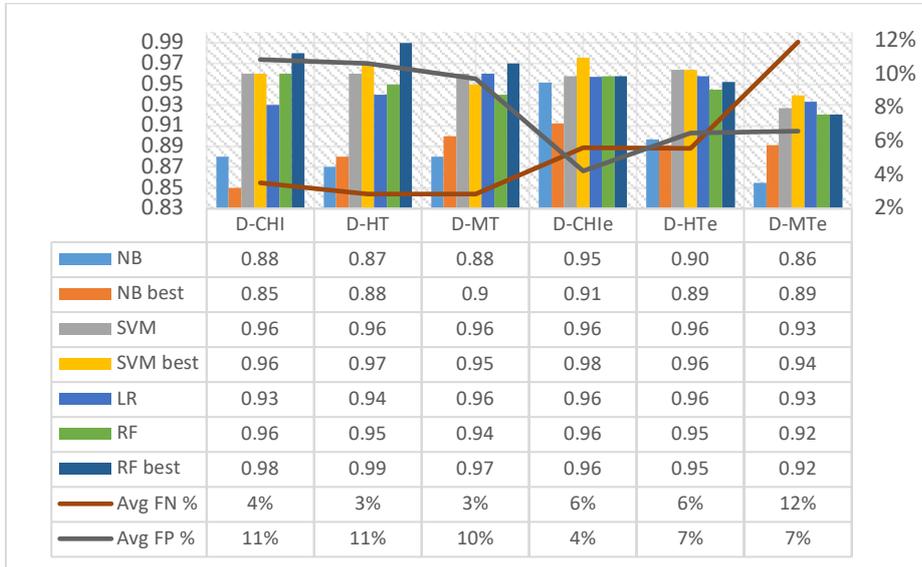

|  | D-CHI | D-HT | D-MT | D-CHIe | D-HTe | D-MTe |
|---|---|---|---|---|---|---|
| NB | 0.88 | 0.87 | 0.88 | 0.95 | 0.90 | 0.86 |
| NB best | 0.85 | 0.88 | 0.9 | 0.91 | 0.89 | 0.89 |
| SVM | 0.96 | 0.96 | 0.96 | 0.96 | 0.96 | 0.93 |
| SVM best | 0.96 | 0.97 | 0.95 | 0.98 | 0.96 | 0.94 |
| LR | 0.93 | 0.94 | 0.96 | 0.96 | 0.96 | 0.93 |
| RF | 0.96 | 0.95 | 0.94 | 0.96 | 0.95 | 0.92 |
| RF best | 0.98 | 0.99 | 0.97 | 0.96 | 0.95 | 0.92 |
| Avg FN % | 4% | 3% | 3% | 6% | 6% | 12% |
| Avg FP % | 11% | 11% | 10% | 4% | 7% | 7% |

**Table 8** lists the FP and FN rates for all models run on the six datasets. Highlighted are rates less than 5%. NB has the worst rates for both FP and FN. Generally, the worse FN rates are on D-MTe.

We can conclude that the addition of more normal SMSs to D-HT and D-MT does not result in better performance and the FN rate deteriorates. As for D-CHI the FP rate improves but the FN rate deteriorates, and the models are worse at detecting fraudulent SMSs. This is an interesting result because it shows that datasets for classifying fraudulent SMSs need not only to be large, but they need to be balanced, and they must contain real-life examples of both fraudulent SMSs and normal SMSs. This may also mean that service-type SMSs need to be classified as their own class rather than being bundled into the normal class.

**A note on removing punctuation and stop-words**
It is often the practice to remove punctuation and stop words in the tokenisation stage for English SMSs. For our Chichewa D-CHI dataset this would require the availability of well-defined stop words for Chichewa, which to our knowledge does not yet exist publicly. Punctuation does not always add important features in general and large texts, however for fraudulent SMSs, punctuation is more relevant and provides important features to classify fraudulent SMSs. To help us understand the impact of removing punctuation and stop-words, we developed a list of stop words from D-CHI. Our experiments showed that removing punctuation and stop words has a negative impact on performance for all models on D-CHI.

The matrix in **Table 8** shows that the number of false negatives on the main dataset D-CHI is low close to 5% except for the models on D-MTe. Models typically aim for a rate of false positives less than 5%. For the translated datasets D-HT and D-MT the rate of false positives and false negatives increases to over 5%. The false positive rate (FP) is much larger than FN rate and this is to be expected given the fact a few words may make all the difference in meaning between similar worded SMSs.

**Precision and Recall**
**Fig. 3** displays accuracy, FR and FP rates for all experiments. All models, except RF best, have a lower precision and a higher recall for the positive class (fraudulent SMS) than for the negative class on all the three datasets. Precision measures the proportion of correctly identified fraudulent SMS among all SMS predicted as fraudulent. Lower precision indicates that models are misclassifying legitimate SMS as fraudulent. Recall measures the proportion of correctly identified fraudulent SMS out of all actual fraudulent **SMS**. This means that models prioritise capturing fraudulent SMS to avoid false negatives, even at the expense of false positives.

In **Table 9** we are conducting a short analysis of misclassified SMSs starting with LR. Messages 1 and 9 are service SMSs sent by telecom provider in Malawi, TNM. LR classifies these as normal (FP). When run on the extended D-CHI these messages are correctly classified correctly as normal. The other messages ( 2-8) have been labelled by the recipient who had direct knowledge of the context. The content of 2 and 4 are too ambiguous for a pure ML system or a human in general to analyse based solely on the text. Similarly message 5 could be fraud depending on circumstances and hence, more context would lead to a better classification. In our experiments all SMSs are considered independent of each other, and we do not analyse conversations, or thread of emails. These examples demonstrate the importance of correct labelling which requires user involvement.

*Table 9 Examples of misclassified SMSs showing the text in Chichewa and translation to English, together with what models misclassify the messages.*

| No | Type | Text in D_CHI | Text in D-HT | Models that misclassify |
|---|---|---|---|---|
| 1 | FP | yankhani mafunso a tnm supa ligi kuti mpeze mwayi okhala m'modzi mwa anthu 4 opata k50,000 sabata iliyonse. tumiza qn ku 451. mtengo ndi k15. 0 1 | Answer TNM Supa League questions to stand a chance of being one of the 4 people to win K50,000 every week. Send QN to 451. The price is K15. | LR, RF, RF best, SVM, SVM best, NB, NB best |
| 2 | FP | nambala yanu siku-mapezeka | Your number has not been available | RF, SVM, SVM best, NB best |
| 3 | FP | kuti tilumikizane imbani foni chifukwa lelo sindibwela | Call for us to communicate because I am not coming today | RF, SVM, SVM best, NB, NB best |
| 4 | FP | ndine deliah | I am Deliah | NB, NB best |
| 5 | FP | watumizidwa katundu wanu akufikani posachdwapa | Your package has been sent, you will receive shortly | NB, NB best |
| 6 | FP | Bwelani ku Maranartha Herbal Healing kuti muzagule mankhwala amatenda osiyanasiyana. | Come to Maranatha Herbal Healing to buy medicine for various diseases | NB, NB best |
| 7 | FN | Social Cash Transfer: banja lanu lakha limodzi mwa maanja opindula ndi social cash transfer. tiyimbileni lamya kuti mulandile ndalama zanu | Social Cash Transfer: Your family has been selected as one of those to benefit with social cash transfer. Call us to receive your money | |
| 8 | FN | okondedwa akasitomala. Dziwani kuti nambala yanu ya chinsisi ibulokedwa pakadutsa 24 hours choncho sinthani nambala yanu ya chinsisi potsitila ndondomeko izi. | Dear customer, Be notified that your secret number will be blocked after 24 hours. Therefore, change your secret number by following these procedures. | RF best, SVM, NB, NB best |
| 9 | FN | Kandalama kaja mungotumizan panumber iyi(0990624230) imalemba martin chimkwita. | If you send the money to this number (0990624230). The name on the account is Martin Chimkwita. | RF, RF best, SVM |

## 4    Discussion and future work

The overall accuracy of our model is comparable to results from similar studies. The benchmark accuracy for the English SMS Spam Collection dataset is 0.97, with SVM being the best-performing model [31]. Machine learning algorithms used to classify SMS messages in non-English languages, such as Indian, Turkish, and Arabic, have also achieved high accuracies, typically exceeding 90% [29], [49], [50], [51]. In contrast, research on using machine learning for SMS fraud classification in African languages remains limited. We identified published studies from Kenya and Ghana [30], [38], [52]. Among these, only one study utilised a dataset in an African language, Swahili [30]. Chichewa, like Swahili, is a Bantu language, hence we compare our results to those obtained for Swahili, and with the benchmark for English in **Table 10**.

Our dataset is smaller in size than the Swahili dataset; however, it contains a larger sample of fraudulent SMS messages—328 compared to 277. The Swahili dataset is highly imbalanced, while our dataset is balanced, providing greater transparency regarding the impact of having a disproportionately large class of normal SMS messages. Our results indicate that as datasets are expanded with more normal SMS messages, the overall performance of the models tends to decrease. A similar observation was made for a large English dataset, highlighting the importance of having a well-curated fraudulent SMS dataset to serve fine-tuning models for one-class classification [42]. For Arabic, a layered approach using NB for an initial classification, followed by a second one-class classifier to correct the fraud predictions has shown improved accuracy [50].

While large datasets are highly desirable for training robust machine learning models, their development comes with significant challenges, the most prominent being accurate labelling. Our analysis revealed instances of SMS messages that were misclassified, highlighting cases where both humans and machines struggle to assign the correct label due to a lack of contextual information. This issue is particularly problematic in scenarios where messages are ambiguous or rely on implicit cultural or situational context. Salman et al. compiled a substantial English SMS dataset containing over 60,000 messages, with approximately one-third labelled as fraudulent [42]. This dataset was built by aggregating messages from various sources and relied on post-collection labelling. However, our findings suggest that the highest accuracy is achieved when SMS messages are labelled directly by their recipients, as this preserves critical contextual insights that are often lost during secondary labelling processes.

Our dataset, though small, exhibits a high degree of diversity compared to other datasets. The English benchmark dataset contains just over 400 unique fraudulent SMS messages, while the Swahili dataset includes only 277. Research has shown that people often fall victim to fraudulent techniques not because they are novel, but because they are repetitive and follow recognisable patterns. Our dataset captures a wide variety of fraudulent SMS messages, offering more diversity in this category than in normal messages. Telecommunication companies send thousands of balance inquiry SMS messages to their users. While these are legitimate and non-fraudulent, they are often considered spam in several datasets we reviewed. Our experiments revealed that adding such normal SMS messages typically degrades the performance of both SVM and RF models, which are otherwise strong performers. These types of normal SMS messages can be automatically labelled and excluded from high-quality SMS datasets used for machine learning. Our findings support a multi-class classification approach, aligning with previous studies that separated SMS categories—such as advertisements or service alerts—from the broader spam class [30], [53].

Table 10 Comparison with other machine learning experiments on English and African Languages SMS datasets

| Similar Research | Almeida et al [31] | Mambina et al [30] | This study |
|---|---|---|---|
| **Country** | USA | Kenya | Malawi |
| **Language** | English | Swahili | Chichewa |
| **Dataset size** | 5,574 | 11,061 | 678 / 824 |
| **Spam (n, % total)** | 747, 13.4% | 277, < 1% | 338, 50% |
| **Accuracy** | 97% (SVM) | 99.86% (RF) | 97% (RF+SVM) |
| **Translations to English** | N/A | None | yes |
| **Classification methods** | SVM, NB, others | RF, NB | RF, NB, SVM, LR |
| **Feature selection** | TF | TF | TF-IDF |
| **Number of features** | 81,175 | 750 | 2000 |
| **Unique SMSs** | Yes | N/A | Yes |
| **Dataset open** | Yes | No | Yes |

**Translation and SMSs in under-resource languages**
It has been noted that models on English datasets that use word embeddings to create features are better at capturing contextual information and relationships between words. Deep learning models that utilise transformers such as BERT are an improvement on shallow deep models and models such as those used in our study [42], [54], [55]. While we used more traditional machine learning techniques (e.g., SVM and RF), incorporating deep learning models like BERT could enhance performance, especially as the size of the dataset grows and the task becomes more intricate [17], [26], [56]. However, transformer-based models require substantial computational resources and larger datasets for training, which might be a challenge in the context of languages like Chichewa, where labelled data is still limited. To our knowledge, there are no embeddings and deep models trained on Chichewa text. To address the limitation of limited labelled datasets for specific languages, some authors have proposed a multi-language approach [25], [36]. This method involves creating datasets that include both English and other languages, leveraging embeddings into English through transfer learning or translation techniques. In this approach, multilingual datasets are created by combining texts from different languages, which are then mapped into a common space (usually English) using pre-trained models or embeddings. Transfer learning allows models trained on large datasets in English to be fine-tuned for other languages, enabling knowledge transfer from high-resource languages to low-resource ones. Alternatively, translation-based approaches convert non-English text into English, allowing the use of existing English-based models without the need for extensive training in the target language.

However, we show that relying on translation into English for either building embeddings for feature construction or for transfer learning has limitations. There is a loss of context during translation that results in a deterioration in performance. Specific terms or phrases important for classification in Chichewa lose their precision or context in English, affecting one class more if it heavily depends on such features. The class fraud may be asymmetrically dependent on specific linguistical patterns, idioms and morphology characteristic to Chichewa. These features may not translate effectively into English, hence both RF and LR performance drops on D-HT and D-MT datasets. Similarly features extracted from D-CHI may not directly map into equivalent features in D-HT and D-MT due to structural and syntactical differences. Random Forest and Logistic Regression are sensitive to the feature distributions they were trained on. If translation alters these distributions (e.g., changes in word frequency or structure), the models' ability to generalize diminishes, particularly for less robust or smaller classes.

Preprocessing methods for text are often understudied and less discussed, yet they are crucial for ensuring that the text is appropriately prepared for machine learning tasks. It is essential that preprocessing techniques be tailored to the specific characteristics of the text type, as the same method that works well for one kind of text (e.g., news articles) may not be suitable for another (e.g., SMS messages). For instance, SMS texts often contain informal language, abbreviations, emojis, and slang, which can pose challenges for traditional preprocessing methods such as stemming or lemmatization. In addition, contextual cues in short messages may be lost during tokenization or stop-word removal if not handled carefully. As seen in our study, certain preprocessing steps, such as full tokenization or removing stop words, led to a deterioration in performance, particularly for Chichewa SMS messages.

Given that text types vary widely across domains, there is a need for domain-specific preprocessing that accounts for the unique linguistic features and contextual dynamics of each text type. For SMS fraud detection, this means considering how fraudulent messages use specific phrasing, urgency, or other behavioural patterns that might not be evident in more formal or larger types of text.

Our main dataset was in Chichewa and the language does not yet have well-developed data preprocessing tools. We developed a small dataset of stop-words for Chichewa derived from the D-CHI dataset, and we observed that performing full tokenization—which involved removing punctuation and stop-words—led to a deterioration in model performance. This finding contrasts with the common approach for English SMSs, where full tokenization is typically effective and often used by default. The performance decline suggests that Chichewa requires language-specific preprocessing to preserve key contextual information during tokenization. This highlights the need for further research into tools tailored for Chichewa and other low-resource African languages, ensuring models capture the full meaning of the text. From a policy perspective, investing in language-specific tools for African languages, especially for applications like SMS fraud detection, is essential to improve model accuracy and effectiveness.

## 5     Conclusion

In this paper, we made several significant contributions with both technical and policy implications. Firstly, we introduced a Chichewa SMS dataset, a valuable resource for one of Africa's widely spoken yet underrepresented languages in the machine learning space. This dataset contains a comparatively large number of fraudulent SMS messages relative to other datasets in the literature. By doing so, we contributed to expanding the availability of SMS spam datasets, which are essential for validating and comparing SMS fraud classifiers across diverse contexts. From a policy perspective, this dataset can serve as a foundation for regulatory bodies and telecommunication authorities in Malawi to promote the design of evidence-based interventions to combat SMS fraud more effectively and to encourage telecommunication companies to invest in such efforts.

Secondly, we conducted an extensive set of machine learning experiments on this dataset, investigating the effects of class balance, translation, and preprocessing techniques on model performance. We compared our results with established benchmarks from English datasets and more recent Swahili datasets, highlighting key differences. Our findings demonstrate that models trained on one language cannot always be directly transferred to another and that preprocessing tools must be adapted to the linguistic nuances of each language. This emphasises the need for language-specific tools tailored to African languages like Chichewa to optimize machine learning outcomes. From a policy standpoint, this insight underscores the importance of investing in language technology research and infrastructure for underrepresented languages to bridge the digital divide.

Using Machine Learning to Detect Fraudulent SMSs in Chichewa    25Thirdly, we proposed a scalable methodology for classifying Chichewa SMS messages, which can be applied to larger datasets in the future. This methodology can guide both academic researchers and industry practitioners in building robust SMS fraud detection systems for Chichewa and related languages. Finally, we provided insights into the key characteristics of fraudulent SMS messages in Malawi, offering a valuable reference for both technical model development and public awareness campaigns. These insights can inform telecommunication regulations, fraud awareness programs, and community outreach initiatives, ultimately contributing to a more resilient communication ecosystem.

Future work will focus on expanding the dataset, exploring advanced data augmentation and feature extraction techniques tailored for Chichewa text, and experimenting with more sophisticated machine learning models, including deep learning architectures. From a policy perspective, collaboration with local stakeholders, such as government agencies and mobile network operators, will be essential to ensure these advancements translate into real-world impact.

**Acknowledgments.** The study did not receive specific funding.

**Disclosure of Interests.** The authors have no competing interests.

**Availability of the data**. The dataset is available on Zenodo https://doi.org/10.5281/zenodo.14607454.## References

[1]   GSMA Intelligence, "The Mobile Economy Sub-Saharan Africa 2024," 2024. Accessed: Jan. 03, 2025. [Online]. Available: https://www.gsma.com/solutions-and-impact/connectivity-for-good/mobile-economy/wp-content/uploads/2024/11/GSMA_ME_SSA_2024_Web.pdf
[2]   Communications Fraud Control Association, "Global Fraud Loss Survey," 2023.
[3]   TrueCalller, "Scam Alert." Accessed: Nov. 22, 2024. [Online]. Available: https://www.truecaller.com/blog/category/scam-alert
[4]   A. Bocereg, M. Tivadar, D. C. Nutiu, and A. Ghinea, "Investigating Worldwide SMS Scams, and Tens of Millions of Dollars in Fraud," Bitdefender Blog. Accessed: Nov. 22, 2024. [Online]. Available: https://www.bitdefender.com/en-us/blog/labs/investigating-worldwide-sms-scams-and-tens-of-millions-of-dollars-in-fraud
[5]   Nation Online, "K120m monthly mobile fraud haunts Macra," The Nation Online, Malawi. Accessed: Nov. 22, 2024. [Online]. Available: https://mwnation.com/k120m-monthly-mobile-fraud-haunts-macra/

28     A. Taylor and A. Robert